\documentclass[letterpaper, 10 pt, conference]{ieeeconf}  
\usepackage{graphicx}
\graphicspath{ {images/} }
\usepackage{amsmath}
\usepackage{xcolor}
\usepackage{floatrow}
\usepackage{subfig}
\usepackage{epstopdf}
\usepackage{hyperref}

\usepackage{algorithm,algpseudocode}
\usepackage{accents}
\usepackage{amsfonts}
\usepackage{wrapfig}
\usepackage{booktabs}
\usepackage{balance}
\usepackage{scalerel}
\makeatletter
\DeclareRobustCommand{\cev}[1]{%
  \mathpalette\do@cev{#1}%
}
\newcommand{\do@cev}[2]{%
  \fix@cev{#1}{+}%
  \reflectbox{$\m@th#1\vec{\reflectbox{$\fix@cev{#1}{-}\m@th#1#2\fix@cev{#1}{+}$}}$}%
  \fix@cev{#1}{-}%
}
\newcommand{\fix@cev}[2]{%
  \ifx#1\displaystyle
    \mkern#23mu
  \else
    \ifx#1\textstyle
      \mkern#23mu
    \else
      \ifx#1\scriptstyle
        \mkern#22mu
      \else
        \mkern#22mu
      \fi
    \fi
  \fi
}

\makeatother
\newcommand{\glassloc}{{\em GlassLoc}}

\long\def\zzm#1{\textcolor{blue}{#1}}

\long\def\ignore#1{ }

\IEEEoverridecommandlockouts                              

\title{\LARGE \bf
GlassLoc: Plenoptic Grasp Pose Detection in Transparent Clutter
}

\author{Zheming Zhou \hspace{0.5cm} Tianyang Pan\hspace{0.5cm} Shiyu Wu\hspace{0.5cm} Haonan Chang\hspace{0.5cm} Odest Chadwicke Jenkins
\thanks{Z. Zhou, T. Pan, S. Wu, H. Chang, and O.C. Jenkins are with the Department of Electrical Engineering and Computer Science, Robotics Institute, University of Michigan, Ann Arbor, MI, USA, 48109-2121  {\tt\small [zhezhou|typan|shiyuwu|harveych|ocj]@umich.edu}}}

\begin{document}


\maketitle
\thispagestyle{empty}
\pagestyle{empty}

\begin{abstract} 
Transparent objects are prevalent across many environments of interest for dexterous robotic manipulation. Such transparent material leads to considerable uncertainty for robot perception and manipulation, and remains an open challenge for robotics.  This problem is exacerbated when multiple transparent objects cluster into piles of clutter. 
In household environments, for example, it is common to encounter piles of glassware in kitchens, dining rooms, and reception areas, which are essentially invisible to modern robots.
We present the {\em GlassLoc} algorithm for grasp pose detection of transparent objects in transparent clutter using plenoptic sensing.  {\em GlassLoc} classifies graspable locations in space informed by a Depth Likelihood Volume (DLV) descriptor.  We extend the DLV to infer the occupancy of transparent objects over a given space from multiple plenoptic viewpoints.
We demonstrate and evaluate the {\em GlassLoc} algorithm on a Michigan Progress Fetch mounted with a first generation Lytro. The effectiveness of our algorithm is evaluated through experiments for grasp detection and execution with a variety of transparent glassware in minor clutter.


\end{abstract}
\section{Introduction}
Robot grasping in household environments is challenging because of sensor uncertainty, scene complexity and actuation imprecision. Recent results suggest that Grasp Pose Detection (GPD) using point cloud local features~\cite{ten2016localizing} and manually labeled grasp confidence~\cite{mahler2017dex} can be applied in generating feasible grasp poses over a wide range of objects.
However, domestic environments include a great amount of transparent objects, ranging from kitchen utilities (e.g. wine cups and containers) to house decoration (e.g. windows and tables). The reflective and transparent material on those objects will produce invalid readings from depth camera.
This problem becomes more significant in the real world where there are piled transparent objects which will lead to unexpected robot manipulation behaviors if the robot was trying to interact with the objects.
A correct estimation of transparency is necessary to protect the robot from performing hazardous actions and extend robot applications to more challenging scenarios. 



The problem of performing grasping in transparent clutter is complicated by the fact that robots cannot perceive and describe the transparent surfaces correctly. Several previous methods~\cite{lysenkov2013recognition,lysenkov2013pose} tried to approach this problem by finding invalid values in depth observation, but they were limited to top-down grasping and made assumption that target objects establish distinguishable contour (formed by invalid points) in depth map. Recently, several approaches employed light field camera to observe the transparency and showed promising results. Zhou et al.~\cite{zhou2018plenoptic} used single shot light field image to form a new plenoptic descriptor named Depth Likelihood Volume (DLV). They succeeded in estimating the pose of single transparent object or object behind translucent surface by given the corresponding object CAD model. Based on that, we extend the idea to a more general-purpose grasp detection scenario with transparent objects clutter.  
\ignore{
In this paper, we propose \glassloc{} algorithm for grasp pose detection in transparent clutter. Based on our extended DLV, \glassloc{} algorithm is capable of detecting multiple grasp poses for the cluttered transparent objects in tabletop settings. For each transparent scene, we construct DLV as the feature of the scene, and generate possible grasp pose samples. Then we use a neural network to perform classification and output the refined grasp poses. The training samples are generated using two kinds of transparent objects, and we test over five kinds of objects cluttered in eight different scene configurations.}

\begin{figure}[t!]
\includegraphics[width=\columnwidth]{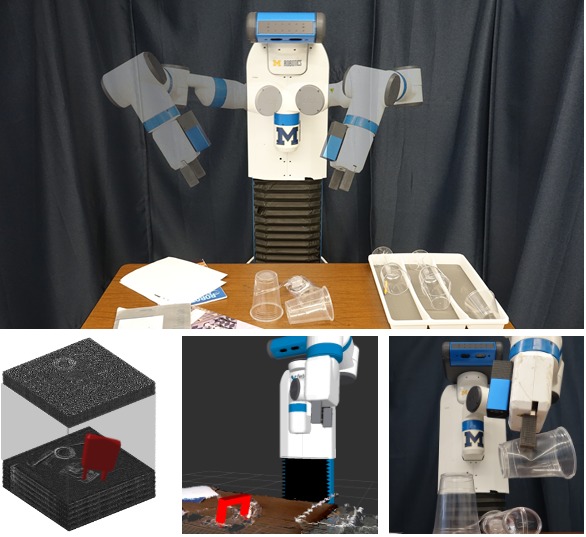}
\caption{(Top) a robot using \glassloc{} to pick up transparent objects from clutter and place on the tray. The robot is observing the scene using a light field camera. Grasp candidate is sampled in DLV (bottom left) and mapped to the world frame in the visualizer (bottom middle). The robot successfully picks up a transparent cup from the clutter (bottom right).
\vspace{-0.5cm}}
\end{figure}
\begin{figure*}[thpb]
\vspace{+1em}
   \centering
      \includegraphics[width=0.95\textwidth]{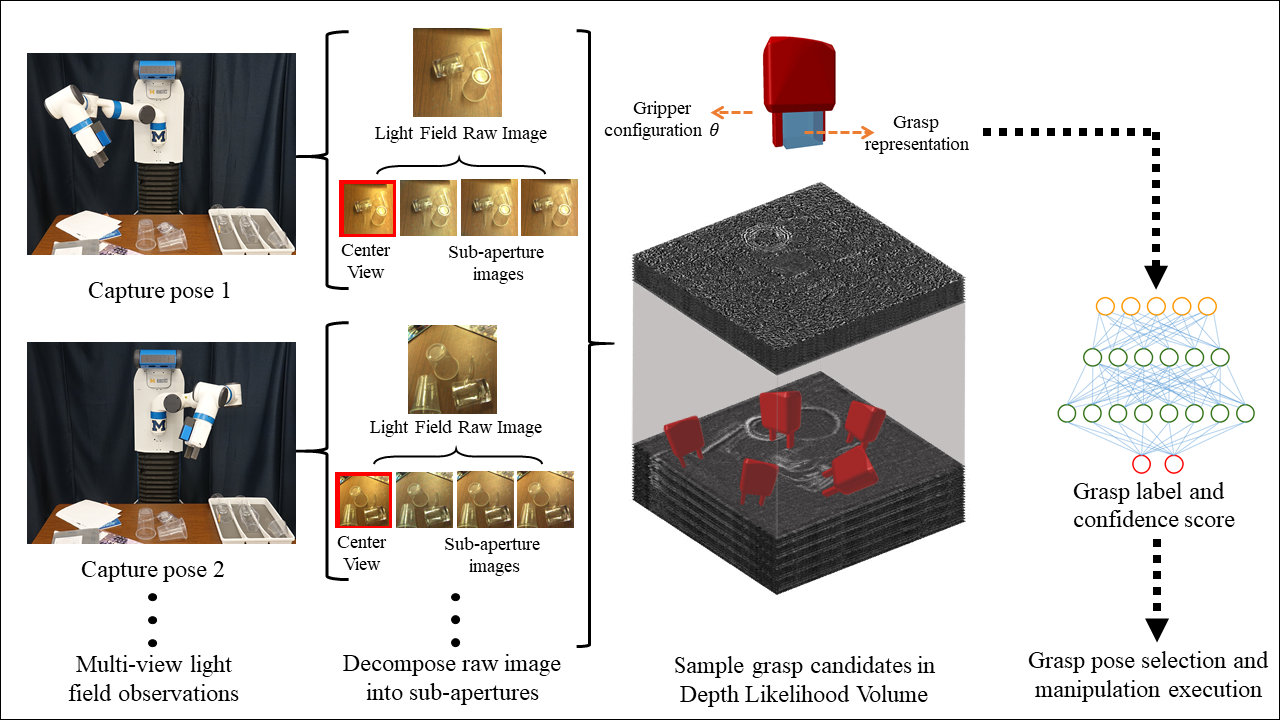}
      \caption{An overview of \glassloc{} framework. A light field camera is mounted on the end-effector of the robot. After taking a set of light field observations by moving robot arms, sub-aperture images are extracted (center view is highlighted in red). The Depth Likelihood Volume (DLV) is then computed as a 3D volume of depth likelihoods over transparent clutter. Given gripper configuration, we can sample grasp poses in DLV and extract grasp features for the classifier to label whether the samples are graspable or not.}
      \label{overview}

\end{figure*}

We make several contributions in this paper. First, we propose \glassloc{} algorithm for detecting six-DoF grasp poses of transparent objects in both separated and minor overlapping cluttered environments. Next, we propose a generalized model for constructing Depth Likelihood Volume from multi-view light field observations with multi-ray fusion and reflection suppression. Finally, we integrate our algorithm with a robot manipulation pipeline to perform tabletop pick and place tasks over eight scenes and five different transparent objects.
Our results show that the grasping success rate over all test objects is 81\% in 220 grasp trials.


\section{Related Work}

\subsection{Grasp Perception In Clutter}
It remains a challenging task for robots to perform perception and manipulation in cluttered environments considering the complexity of the real world. We consider there are two major categories of methods for robots to perform grasp perception in clutter. The first category is model-based pose estimation methods. By estimating object poses, grasp configurations calculated based on the local model can be further transformed to the robot environments.
Collet et al.~\cite{MOPED} utilized color information to estimate poses of object in cluttered environments. Their proposed algorithm clusters and then matches the local color patch from object model to robot observations to generate pose hypotheses. 
Sui et al.~\cite{Suietal_SUM,Suietal2017ijrr} constructed generative models to evaluate pose hypotheses against point cloud using object CAD models. The generative models perform object detection followed by particle filtering for robot grasping in the highly cluttered tabletop environments. 
With a similar idea, Papazov et al.~\cite{papazov2012rigid} leveraged RANSAC-based bottom-up approach with Iterative Closest Point registration to fit 3D geometries to the observed point cloud.

On the other hand, rather than associating a grasp pose with a certain object model, Grasp Pose Detection (GPD) tries to characterize grasp poses based on the local geometry or appearance features directly from observations. 
Several early works~\cite{lenz2015deep,redmon2015real} represented the grasp poses as oriented rectangles in RGB-D observations. Further, given a number of manually-labelled grasp candidates, the system will learn to predict whether a sampled rectangle is graspable or not. One major restriction of those systems is that the approaching directions of generated grasp candidates need to be orthogonal to the RGB-D sensor plane. Fischinger and Vincze~\cite{fischinger2012empty} tried to lessen the restriction by integrating hightmap-based features. They also designed a heuristic for ranking the grasp candidates in a clutter bin settings. ten Pas and Platt~\cite{tenPas2015} directly detected grasp poses in $SE(3)$ space by estimating curvatures and extracting handle-like features in local point cloud neighborhoods. Gualtieri et al. \cite{gualtieri2016high} proposed more types of local point cloud features for grasp representation and projected those features to 2D image space for classification.
Our work with \glassloc{} extends these ideas to transparent clutter with a different grasp representation and a new plenoptic descriptor.

\subsection{Light Field Photography}
The models describing the light field rendering proposed by Levoy and Hanrahan \cite{levoy1996light} introduced foundations of light field captured from multi-view cameras. Based on this work,~\cite{ng2006digital,georgiev2013lytro} succeeded in producing commercial level hand-held light field camera using the microlens array structure. Building on the property that the plenoptic camera can capture both intensity and direction of light rays, light field photography has shown significant advancement in different applications. Wang et al.~\cite{wang2015occlusion} explicitly modeled the light field image pixel angular consistency to generate accurate depth map for the object with occlusion edges. Jeon et al.~\cite{jeon2015accurate} performed sub-pixel shifting in image frequency domain in tackling the microlens camera narrow baseline problem for accurate depth estimation. Maeno et al.~\cite{maeno2013light} introduced distortion feature in light field to detect and recognize the transparent object. Johannsen et al.~\cite{reconst_accv2017} leveraged multi-view light field images to reconstruct multi-layer translucent scenes. Skinner and Johnson-Roberson~\cite{skinner2016towards} introduced a light propagation model suited to underwater perception using plenoptic observations.

The use of light field perception in robotics is still relatively new. Oberlin and Tellex~\cite{oberlintime} proposed a time-lapse light field capturing pipeline for static scenes by mounting a RGB camera on the end-effector of the robot and moving in a designed trajectory. Tsai et al.~\cite{tsai2018distinguishing} introduced a algorithm for distinguishing refracted and Lambertian features from light field image. Zhou et al.~\cite{zhou2018plenoptic} used a Lytro camera to take a single shot of the scene and construct a plenoptic descriptor over that. Given the target object model, their methods can estimate single object six-DoF pose in layered translucent scenes. Our \glassloc{} pipeline extends the idea proposed in~\cite{zhou2018plenoptic} for more general-purpose manipulation over transparent clutter.

\section{Problem Formulation and Approach}

{\em GlassLoc} addresses the problem of grasp pose detection for transparent objects in clutter from plenoptic observations.
For a given static scene, we assume there is a latent set of end-effector poses $G \subset SE(3)$ that will produce a successful grasp of an object.  A successful grasp is assumed to result in the robot obtaining force closure on an object when it moves gripper and closes its fingers.   The plenoptic grasp pose detection problem is then phrased as estimating a representative set of valid sample grasp poses $G_\textrm{v} \subset G$.  

Within the grasp pose detection problem, a major challenge is how to classify whether a grasp pose is a member of $G$, and, thus, will result in a successful manipulation.  For grasp pose classification, we assume as given robot end-effector pose $q \in SE(3)$ and a collection of observations $Z$ from a plenoptic sensor. It is assumed that each observation $z_{1:N} \in Z$ captures a raw light field image $o_i$ of a static scene from camera viewpoint $v_i \subset SE(3)$.  
The classification result calculated from these inputs is a likelihood $l \in [0,1]$ that relates the probability of end-effector pose, $q$, resulting in a successful grasp.  Described later, our implementation of {\em GlassLoc} will perform the classification using a neural network.

\begin{figure}[t!]
\includegraphics[width=\columnwidth]{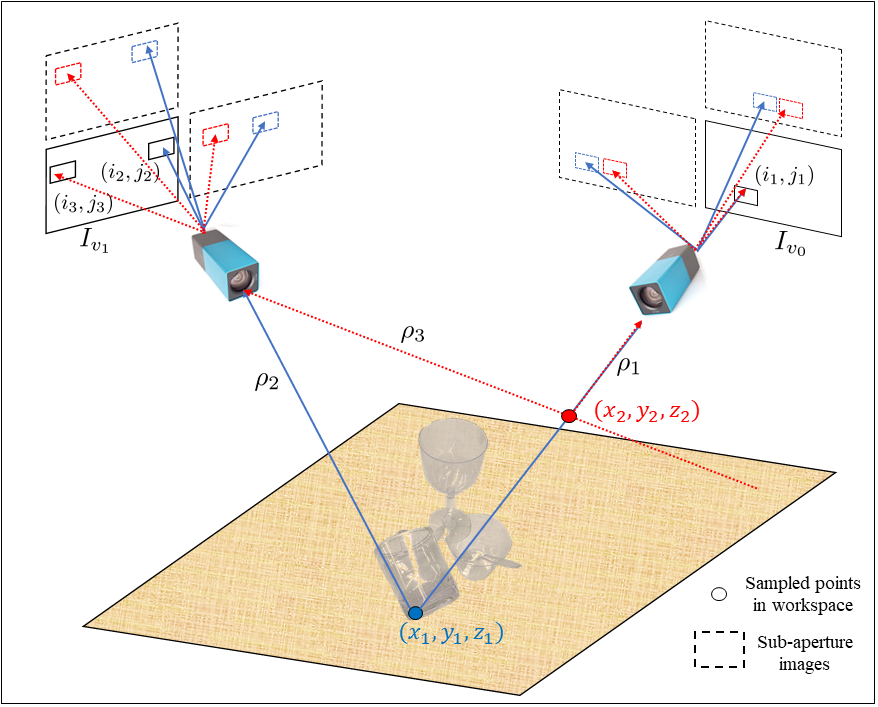}
\caption{Example of DLV value calculation of two randomly sampled points $(x_1,y_1,z_1)$ and $(x_2,y_2,z_2)$ through examining the ray consistency in different view points. Each sample point corresponds to different pixel indices with depths in different center view plane $I_{v_0}$ and $I_{v_1}$. 
}
      \label{fig:explain_dlv}
\end{figure}

Illustrated in Figure~\ref{fig:explain_dlv}, grasp pose classification within {\em GlassLoc} is expressed as a function $l = \mathcal{M}(U)$  
that maps transparency occupancy likelihood features $U$ to grasp pose confidence $l$. Transparency occupancy features $U(q,D)$ are computed with respect to the subset of a Depth Likelihood Volume (DLV) $D$ that is within the graspable volume of pose $q$.  The DLV estimates how likely a point $p \in \mathbb{R}^3$ belongs to a transparent surface.  To test all sampled grasps, a Depth Likelihood Volume $D$ is computed from observations $Z$ over an entire grasping workspace $P \subset SE(3)$ within the visual hull of $v_{1:N}$.  We assume the grasping workspace is discretized into $p_{1:M} \in P$ a set of 3D points, with each element of this set expressed as $p_i = (x_i,y_i,z_i)$.

\section{Plenoptic Grasp Pose Detection Methods}
An outline of the \glassloc{} algorithm is described in Algorithm~\ref{alg:PGPD}.  \glassloc{} begins by computing a Depth Likelihood Volume from multi-view light field observations. By integrating different views, we can further post-process the DLV by suppressing reflection caused by non-Lambertian surfaces. Details of DLV construction are presented in Section~\ref{subsec:dlv} and~\ref{subsec:rs}.
In Step 2, we uniformly sample the grasp candidates $C = \{c_j\in P\}$ 
in workspace $P$. For each grasp candidate, we extract grasp representations 
(see Section~\ref{subsec:grc}) and corresponding transparency likelihood features 
given the robot gripper parameter $\theta$.
The generated features will then be classified with a grasp success labels and confidence scores by a neural network. The training data generation strategy for learning this mapping is introduced in Section~\ref{subsec:tdg}.
Given classified grasp poses, we use a multi-hypothesis particle-based search
to find a set of end-effector poses with high confidence for successful grasp execution (see Section~\ref{subsec:gf}). The finalized set of grasp poses will be ready for the robot to perform grasping.

\begin{algorithm}[!t]
  \caption{\glassloc{} Plenoptic Grasp Pose Detection}
  \label{alg:PGPD}
  \begin{flushleft}
    \textbf{INPUT:} a set of light field observations $Z$, robot gripper parameter $\theta$\\
    \textbf{OUTPUT:} a set of valid sample grasp poses $G_v$
  \end{flushleft}
  \begin{algorithmic}[1]
   \State $D = \text{Construct\_DLV(Z)}$
   \State $C = \text{Sample\_Grasp\_Candidates}(D,\theta)$
   \For{$i = 1 ... K$}
   \State $G_{i} = \text{Grasp\_Classification(}C\text{)}$
   \State $C = \text{Resample\_Diffuse}(G_{i})$
   \EndFor
   \State $G_{\text{v}} \gets C$
  \end{algorithmic}
\end{algorithm}

\subsection{Multi-view Depth Likelihood Volume}
\label{subsec:dlv}

\ignore{
\begin{figure*}[!t]
\vspace{1em}
   \centering
      \includegraphics[width=0.35\textwidth]{figure/multi_view_general_crop.jpg}
      \includegraphics[width=0.620\textwidth]{figure/multi_view_detail.jpg}
      \caption{(Left) a scene with a transparent clutter contains four transparent instances. Multi-view light field capture illustration (here we take two camera poses as example). (Right) Illustration of DLV step. The depth values sampled from reference camera pose are transformed to second view to find the corresponding pixel index.}

      \label{depth_dist}
\end{figure*}
}
The Depth Likelihood Volume (DLV) is a volume-based plenoptic descriptor which represents the depth of a light field image pixel as a likelihood function rather than a deterministic value. The advantage of this representation is to keep the transparent scene structure by assigning different likelihoods to surfaces with different transparency. In~\cite{zhou2018plenoptic}, DLV is formulated in a specific camera frame indexed with pixel coordinates and depths. The formulation is restricted to single-view scenarios.
In this paper, we generalize the expression which takes sample points in 3-D space as input and integrates multi-view light field observations. 

The DLV is defined as:



\ignore{
\begin{equation}
\label{eq:mdlv}
L(x,y,z) =
\sum_{\substack{\rho_{\{i,j\}}\\
\in\cev{\rho}_{\{x,y,z\}}}}\sum_n\frac{\max_k T_n^k{(\rho_{\{i,j\}})} - T_n^d{(\rho_{\{i,j\}})}}{\sum_k T_n^k{(\rho_{\{i,j\}})}}
\end{equation}
}

\begin{equation}
\label{eq:dlv}
L(p) =
\sum_{i}^Nf\bigg(\sum_{\scaleto{a\in A\backslash I_{v_i}}{6pt}}T_{a,d}(\rho_{v_i}(p))\bigg)
\end{equation}
\begin{equation}
\label{eq:t}
T_{a,d}{(\rho)} = ||\rho,\mathcal{F}_{a,d}(\rho)||
\end{equation}

where $L(p)$ is the depth likelihood of sampled points $p$. $A$ is the set of sub-aperture images. $\rho_{v_i}(p)$ is a light ray that goes through or emitted from point $p$ and is received by view point $v_i$ at $(i,j)$ in center view image plane. $N$ indicates the number of view points in observations. $\mathcal{F}_{a,d}(\rho)$ is the triangulation  function finding the light ray corresponding to $\rho$ in sub-aperture images indexed with $a$ that yields depth $d$. $d$ can be explicitly calculated using camera intrinsic matrix given point and view point. $||\cdot,\cdot||$ is the ray difference which is calculated by color and color gradient differences. Denote $s=\sum_{\scaleto{a\in A\backslash I_{v_i}}{6pt}}T_{a,d}(\rho_{v_i}(p))$, then $f(s)$ is a normalization function mapping color cost to likelihood. There are multiple choices of $f(s)$. In our implementation, we choose:
\begin{equation}
\label{eq:f}
f(s) = 
\frac{\max_k\sum_{\scaleto{a\in A\backslash I_{v_i}}{6pt}}T_{a,k}(\rho_{v_i}(p)) - s}{\sum_k\sum_{\scaleto{a\in A\backslash I_{v_i}}{6pt}}T_{a,k}(\rho_{v_i}(p))}
\end{equation}
\ignore{
Notice that, the $\rho$ in Equation~\ref{eq:f} must be the ray received by the same pixel $(i,j)$ indicated in Equation~\ref{eq:dlv}. }

To better explain the formulation presented above, we consider the example shown in Figure~\ref{fig:explain_dlv}. A cluster of transparent objects are placed on a table with opaque surface. We have two light field observations $z_0 = \{o_0,v_0\}$ and $z_1 = \{o_1,v_1\}$ with center view image plane $I_{v_0}$ and $I_{v_1}$ respectively. There are two points $p_1=(x_1,y_1,z_1)$ and $p_2= (x_2,y_2,z_2)$ sampled in the space and each of them emits light rays captured by both views. In view $I_{v_0}$, Ray $\rho_1$ emitted from both points are received by the same pixel $(i_1,j_1)$, while $\rho_2$ and $\rho_3$ are received by $(i_2,j_2)$ and $(i_3,j_3)$ respectively. Then we can express the depth likelihood of point $p_2$ as:
\begin{equation}
    \label{eq:example_2L}
    L(p_2) = f\bigg(\sum_{\scaleto{a\in A\backslash I_{v_0}}{5.8pt}}T_{a,d_1}(\rho_1)\bigg)
     + f\bigg(\sum_{\scaleto{a\in A\backslash I_{v_1}}{5.8pt}}T_{a,d_3}(\rho_3)\bigg)
\end{equation}
\ignore{The normalization function $f$ for the first term can be further expanded according to Equation~\ref{eq:f}:
\begin{equation}
    \label{eq:example_2f}
    \frac{\max\{\sum_nT^{d}_n(\rho_1),\sum_nT^{d}_n(\rho_2)\} - \sum_nT^{d}_n(\rho_1)}{\sum_nT^{d}_n(\rho_1) + \sum_nT^{d}_n(\rho_2)} 
\end{equation}}
Function $T$ calculates the color and the color gradient difference between center view (rectangle with solid line in Figure~\ref{fig:explain_dlv}) and sub-aperture view (rectangle with dot line in Figure~\ref{fig:explain_dlv}). The location of red pixel is calculated by function $\mathcal{F}$. For micro-lens based light field camera, the pixel shift between center and sub-aperture images are usually in sub-pixel level. The realization of $\mathcal{F}$ function is based on frequency domain sub-pixel shifting method proposed in~\cite{jeon2015accurate}.

\subsection{Reflection Suppression}
\label{subsec:rs}

\begin{figure}[t!]
\includegraphics[width=\columnwidth]{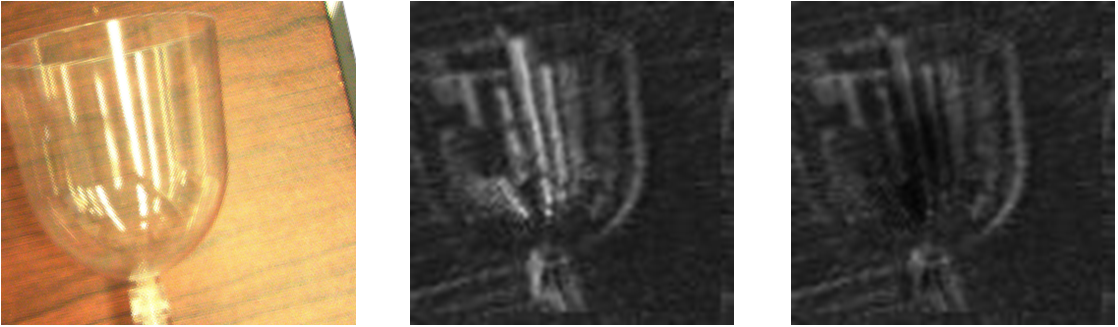}
\caption{Example DLV feature image before (middle) and after (right) reflection suppression. The center view of part of raw observation is shown in (left). The intensity of pixel in the gray-scale image (middle and right) indicates the likelihood value. The high likelihood region caused by specular light is suppressed.
}
      \label{fig:reflective_s}
\end{figure}



A transparent surface produces non-Lambertian reflectance, which induces specular highlight to light field observations. Those shiny spots tend to produce the saturated color or virtual surface with larger depth than the actual transparent surface. This phenomenon will generate a high likelihood region in DLV that indicates a non-existing surface. To deal with this problem, we calculate the variance of ray differences for DLV points which has saturated color and high likelihood over different view points:
\begin{equation}
   var\{\rho_V(p)\} = \sum_i^N\sum_{\scaleto{a\in A\backslash I_{v_i}}{6pt}}(T_{a,d}(\rho_{v_i}(p))- E\{\rho_V(p)\})^2
\end{equation}
where $E\{\rho_V(p)\}$ can be expressed as:
\begin{equation}
   E\{\rho_V(p)\} = \frac{1}{N\cdot (N(A)-1)} \sum_i^N\sum_{\scaleto{a\in A\backslash I_{v_i}}{6pt}}T_{a,d}(\rho_{v_i}(p))
\end{equation}
where $N(A)$ is the number of sub-aperture images extracted from raw light field image.
For a point $p$ that has variance larger than a threshold $\tau$, we check whether it has the largest likelihood value among all other points that lie on the light rays it emits out. Specifically, we first find light rays emitted from $p$ and received by pixel $(i,j)$ with depth $d$ that has large variance over different view points. Then we locate all light rays received by $(i,j)$ with depth less than $d$, and check whether the following equation holds:
\begin{equation}
\label{eq:reflective}
    \max_k\sum_{\scaleto{a\in A\backslash I_{v_i}}{6pt}}T_{a,k}(\rho_{v_i}(p)) = \sum_{\scaleto{a\in A\backslash I_{v_i}}{6pt}}T_{a,d}(\rho_{v_i}(p))
\end{equation}
If Equation~\ref{eq:reflective} holds, it indicates this light ray has high possibility of coming from strong reflection area and will be excluded from the calculation of DLV. Figure~\ref{fig:reflective_s} (left) is the sliced feature from DLV before reflective suppression which we can observe incorrect large values caused by specular highlight. Figure~\ref{fig:reflective_s} (right) shows the result after processing and the previous high value area is suppressed.

\begin{figure}[t!]
\subfloat[\footnotesize]{
\includegraphics[width=0.46\linewidth]{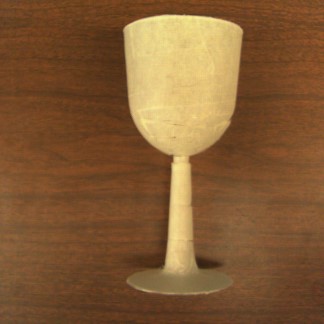}
}
\subfloat[\footnotesize]{
\includegraphics[width=0.46\linewidth]{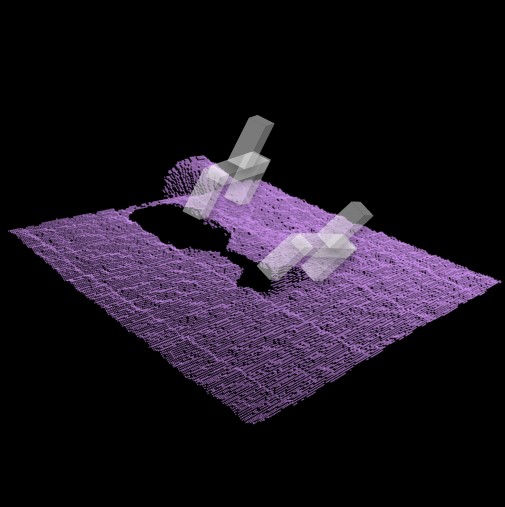}
}

\subfloat[\footnotesize]{
\includegraphics[width=0.46\linewidth]{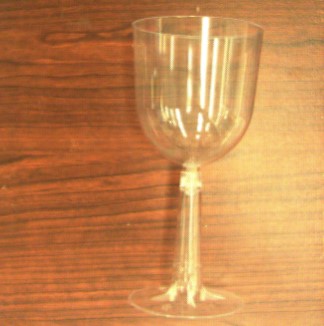}
}
\subfloat[\footnotesize]{
\includegraphics[width=0.46\linewidth]{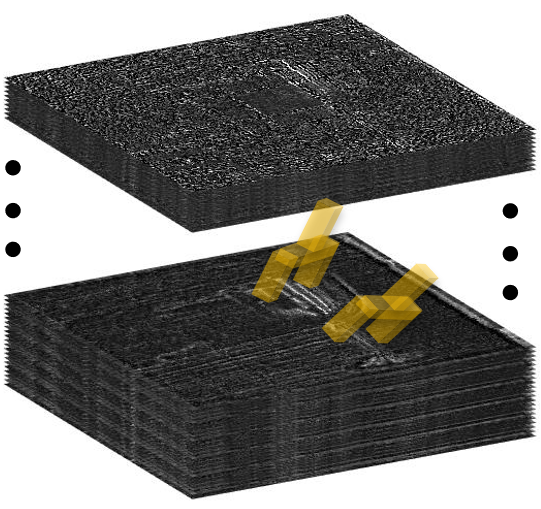}
}
\caption{Training data generation procedure. (a) The glass cup is wrapped with opaque tape for depth sensor to get point cloud. (b) Grasp candidates are generated based on point-cloud-based method and local-to-world transform. (c) The glass cup is placed at the same pose to take multiple light field observations. (d) Grasp candidates generated from point cloud are mapped to DLV.}
\label{fig:training_steps}
\end{figure}

\subsection{Grasp Representation and Classification }
\label{subsec:grc}
We represent a graspable area as a 3D cuboid with length, width, and height as $L,W,H$ respectively. 
The width and height of the cuboid is equal to the width and height of the volume when the robot finger close while the length is extended for capturing more feature spaces. 
The cuboid is voxelized into $l\times w\times h$ grid, and for each grid we interpolate the likelihood value by finding the nearest eight points in DLV. Rather than feeding into classifier with a large amount of points, we extract 2D features from the volume by projection and slicing.

We first define the three axes of the graspable volume. The $x$ axis of the volume is defined as the approach direction of the gripper. The $z$ axis is defined along the direction the gripper fingers close along. The $y$ axis is the cross product of the previous two axes. We then calculate three types of features and project them to the three axes: a center slice of likelihood volume, $I_c$, an average likelihood map over all points, $I_a$, a sliced difference likelihood map, $I_d$, which is calculated by recursively comparing the difference between current slice of the graspable volume with the previous slice. More specifically, we can express the three types of feature as follows (take projection to $x$ axis as example):
\begin{equation}
\label{eq:grasp_rep1}
    I_c(x,y) = L(x,y,z=\frac{h}{2}) 
\end{equation}
\begin{equation}
\label{eq:grasp_rep2}
    I_a(x,y) = \frac{\sum_{z=0}^hL(x,y,z)}{h}
\end{equation}
\begin{equation}
\label{eq:grasp_rep3}
    I_d(x,y) = \frac{\sum_{z=0}^{h-1}|L(x,y,z)-L(x,y,z+1)|}{h} 
\end{equation}
We resize the images to the same size and concatenate them into different channels. Since we have three types of features and three axes to project, we have nine channels in total. 

For classifier, we use the LeNet~\cite{lecun1998gradient} structure which is a common structure for grasp pose classification and ranking~\cite{gualtieri2016high,kappler2015leveraging}. The output of the classifier is the binary label $\{graspable, not\;graspable\}$ associated with the confidence scores. 

\ignore{\zzm{Notice that, considering the nature 3D structure of the mDLV, the grasp representation is compatible with other 2D or 3D-based neural network.}}

\subsection{Training Data Generation}
\label{subsec:tdg}

For depth-based grasp pose detection algorithms, the training data generation process relies on grasp pose sampling and labeling on point cloud. Unfortunately, depth sensors cannot provide correct point cloud for transparent objects. Instead, we wrap the object with opaque material and generate training samples by mapping grasp poses from point cloud to DLV. The detailed steps are illustrated in Figure~\ref{fig:training_steps} (a) - (d).

We have two sources to produce training samples from point cloud. One is depth-based grasp pose detection algorithms. We input those algorithms with our depth observations and label the result grasp candidates as $\{graspable\}$. In the meantime, we restore the grasp poses filtered out in those algorithms and label them as $\{not\;graspable\}$. The other is transforming pre-defined grasp pose in the local frame to the observation. By checking the gripper collision with the environment, we label the collision free grasp poses as $\{graspable\}$ and the others as $\{not\;graspable\}$. 

\begin{figure}[t!]
\includegraphics[width=\columnwidth]{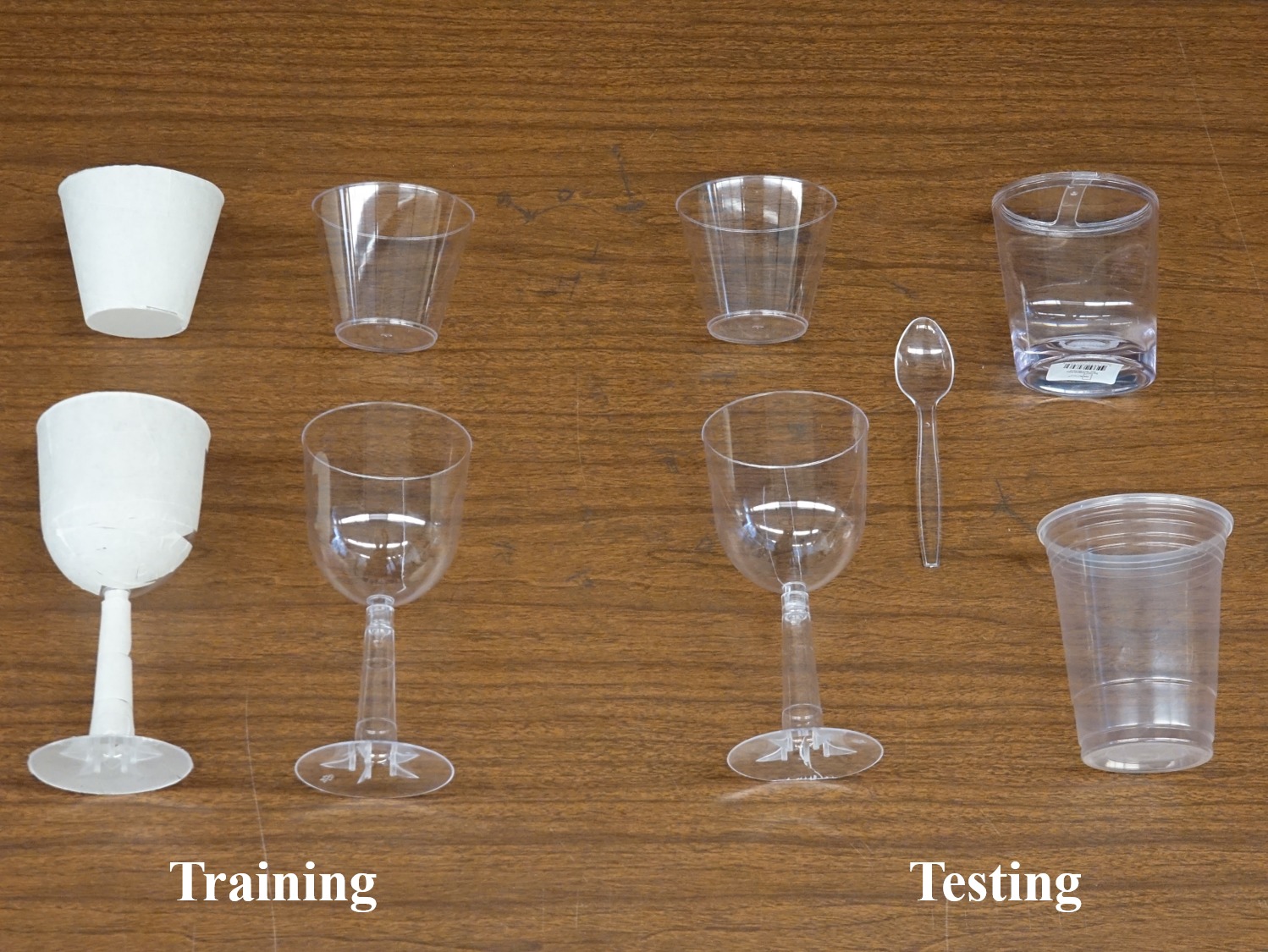}
\caption{Training and testing objects for evaluating our \glassloc{} algorithm. Two objects are used in training: wine cup and short cup (wrapped object for generating point cloud). Five objects are used in testing: wine cup, toothbrush holder, spoon, short cup, and tall cup. 
}
\label{fig:train_test}
\end{figure}


\begin{figure*}[t!]
\subfloat[\footnotesize]{
\includegraphics[width=0.23\linewidth]{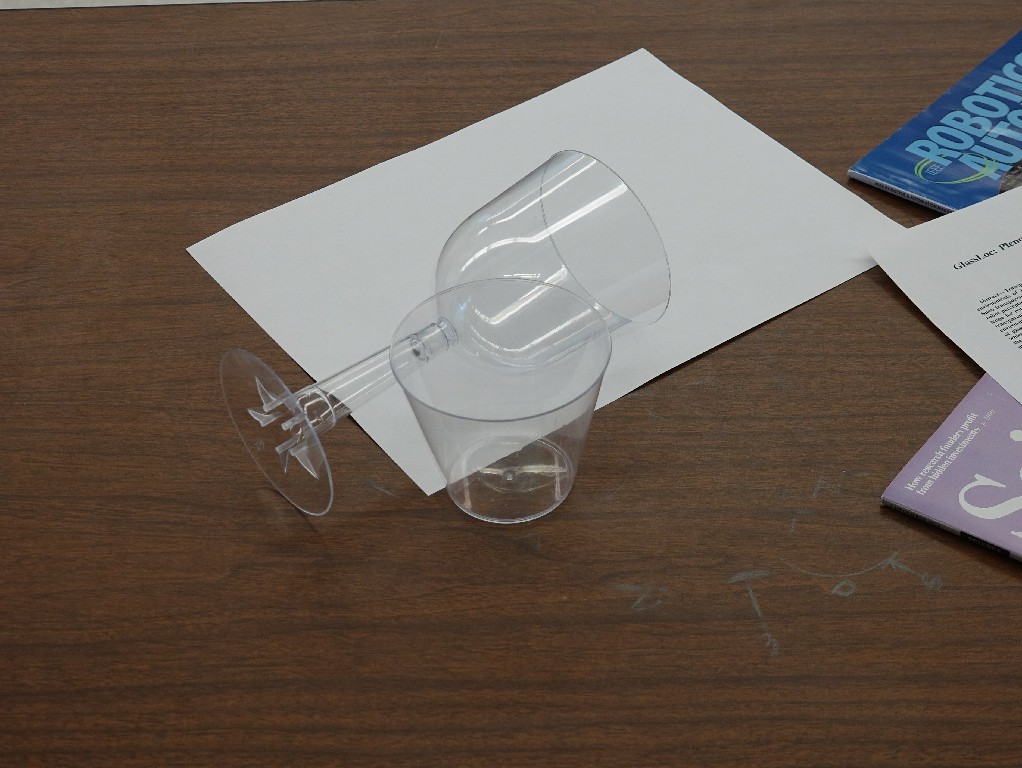}
}
\subfloat[\footnotesize]{
\includegraphics[width=0.23\linewidth]{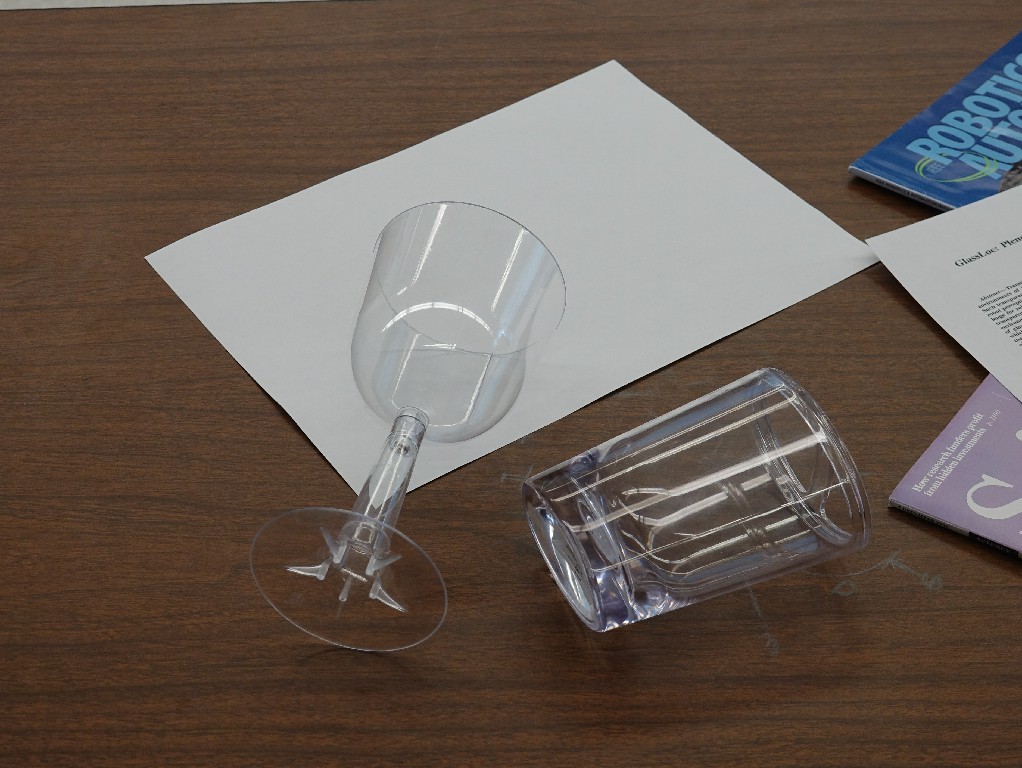}
}
\subfloat[\footnotesize]{
\includegraphics[width=0.23\linewidth]{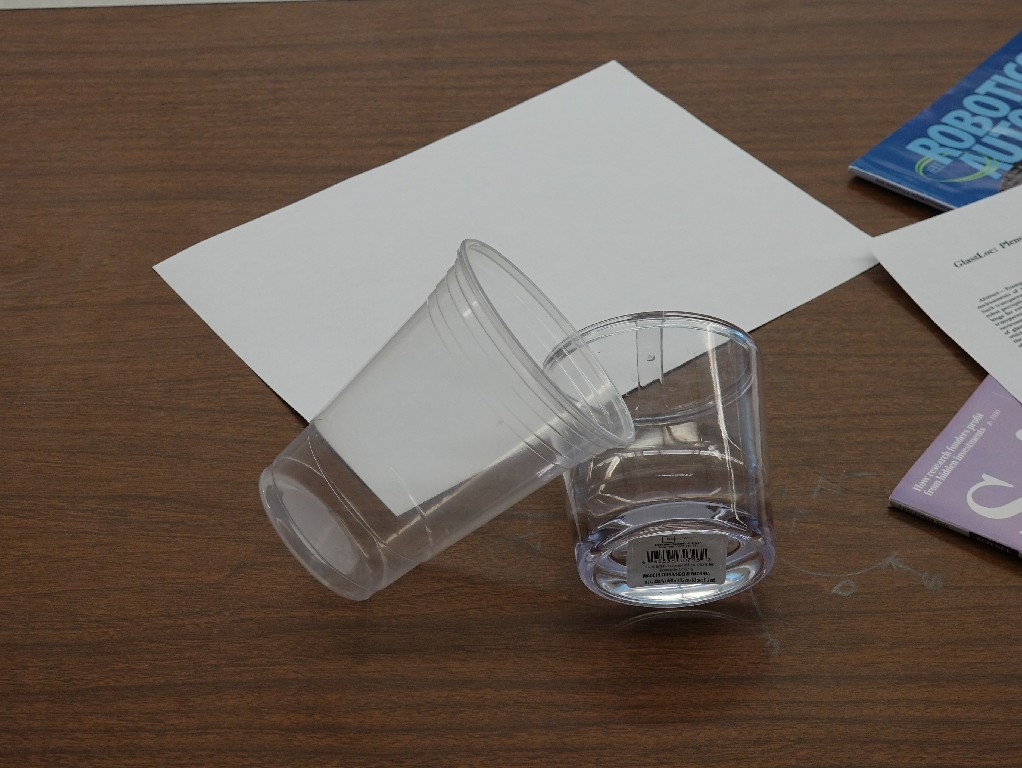}
}
\subfloat[\footnotesize]{
\includegraphics[width=0.23\linewidth]{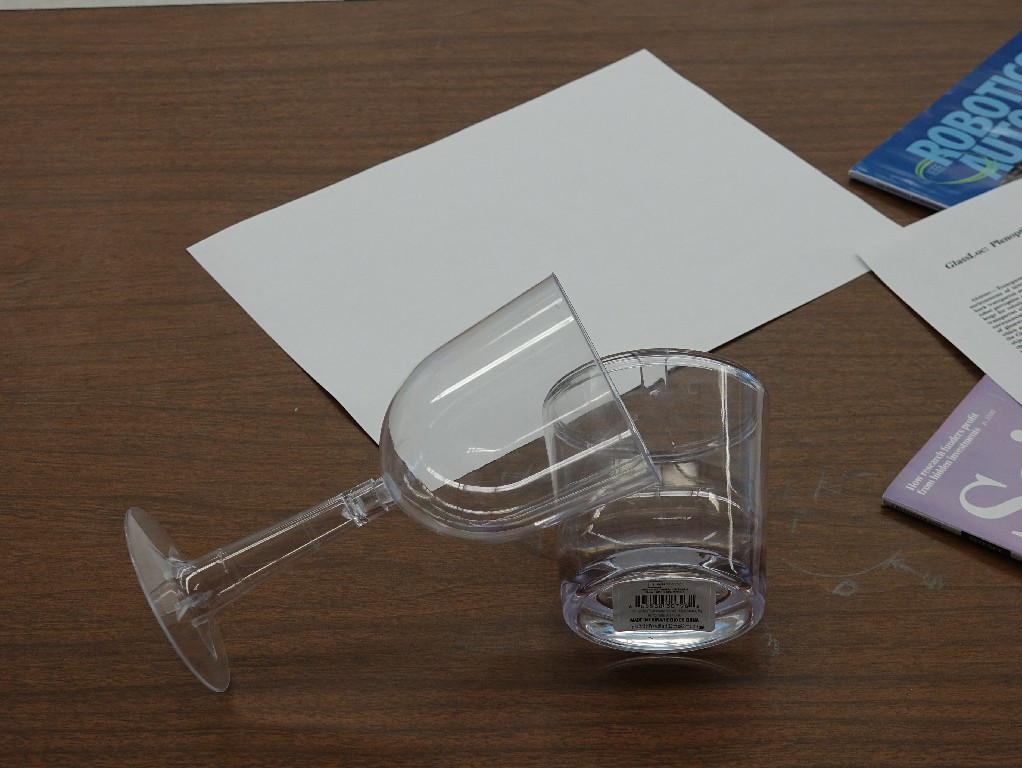}
}

\subfloat[\footnotesize]{
\includegraphics[width=0.23\linewidth]{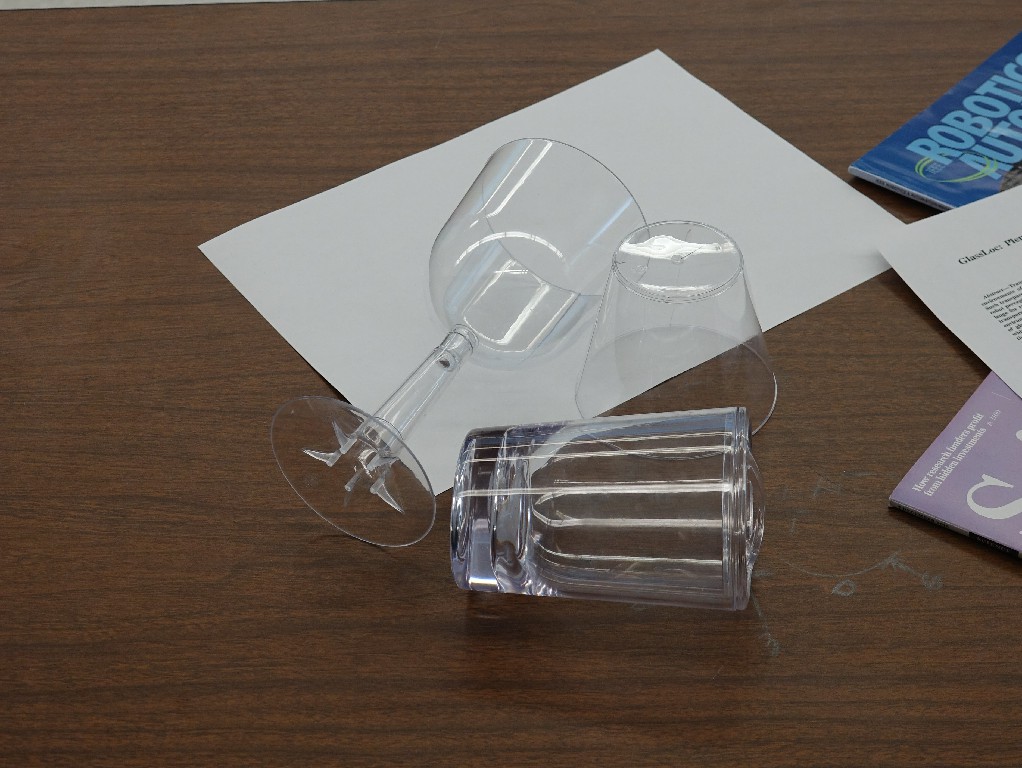}
}
\subfloat[\footnotesize]{
\includegraphics[width=0.23\linewidth]{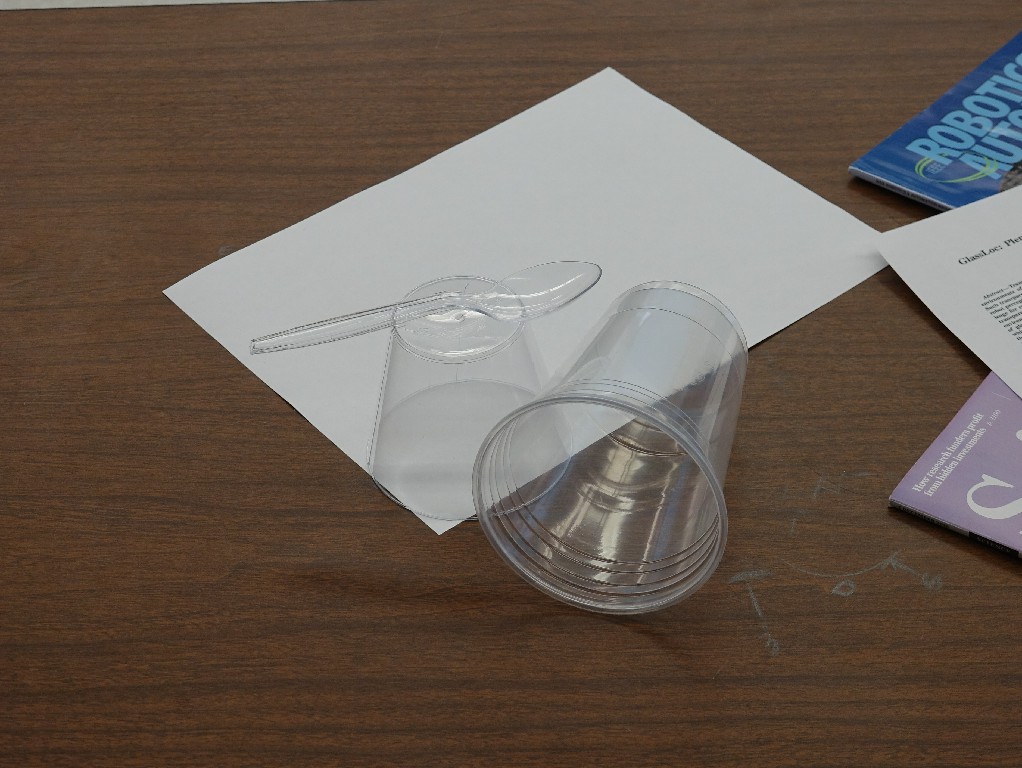}
}
\subfloat[\footnotesize]{
\includegraphics[width=0.23\linewidth]{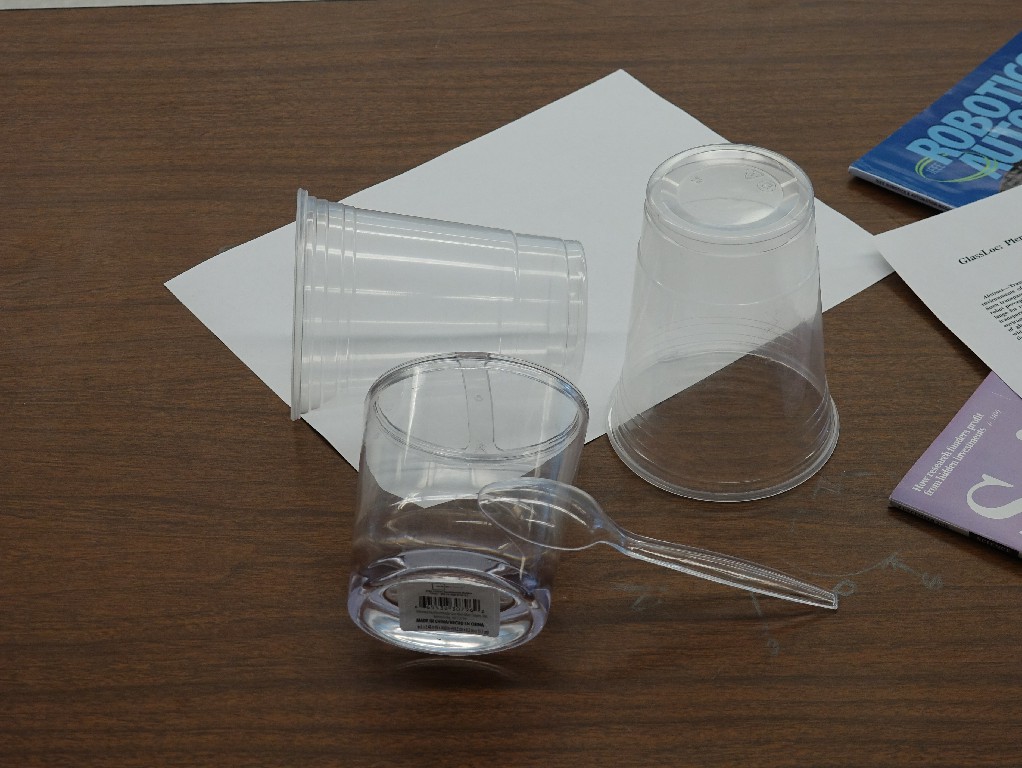}
}
\subfloat[\footnotesize]{
\includegraphics[width=0.23\linewidth]{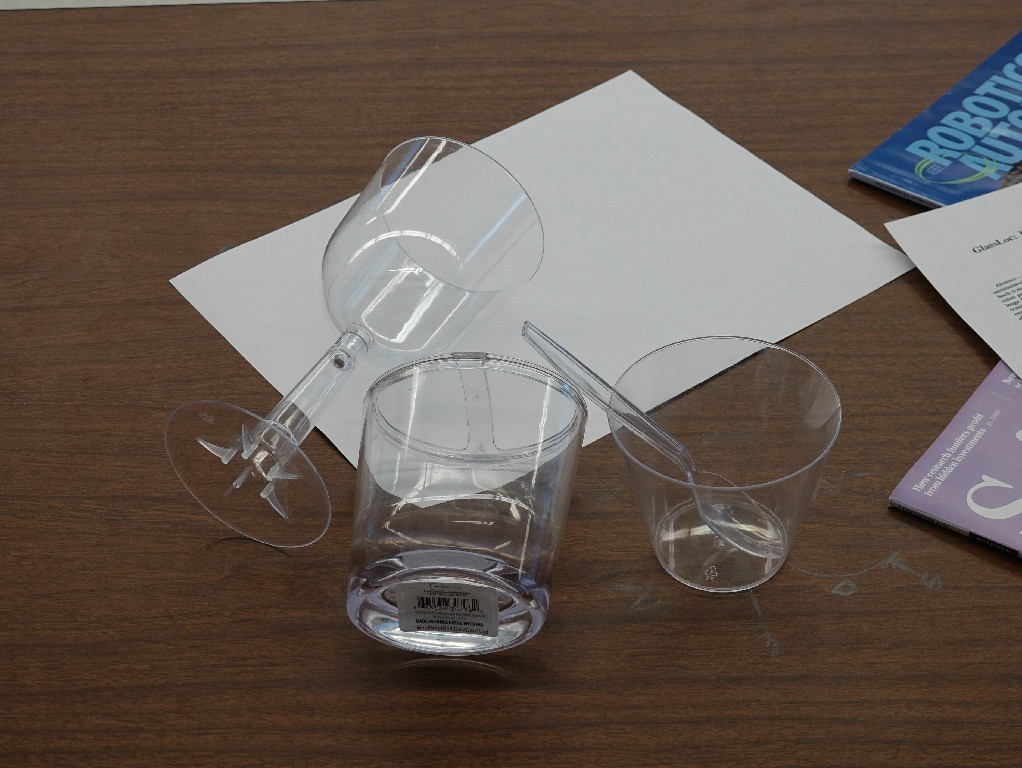}
}

\caption{Eight scenes for evaluating \glassloc{} pipeline. We randomly choose a number of transparent objects from the test set and put them on the table for the robot to perform manipulation on.}
\label{fig:test_scenes}
\end{figure*}
\begin{table*}
\tiny
\centering
\resizebox{\columnwidth}{!}{
\begin{tabular}{c@{\quad}ccccccccc}
 \hline
 & scene (a) & scene (b) & scene (c) & scene (d) & scene (e) & scene (f) & scene (g) & scene (h) \\
  \hline
   Number of \\ Total Objects & 2 & 2 & 2 & 2 & 3 & 3 & 4 & 4 \\[0.05cm]
 		
  Number of \\ Manipulation Runs & 10 & 10 & 10 & 10 &  10 & 10 & 10 & 10 \\[0.05cm]
  	
  Object Grasp \\ Percentage & 0.70 & 0.80 & 1.0 & 0.75  & 0.87 & 0.43 & 1.0 & 0.85 \\[0.05cm]
   \hline
\end{tabular}
}
\caption{Results of manipulation experiments for eight scenes. The first row shows the number of object in the scene. Number of manipulation runs shown in row two refers to the task runs for the scene. The object grasp percentage refers to successful picking ratio over all trials for each scene.}
\label{table:results}
\end{table*}

\subsection{Grasp Search}
\label{subsec:gf}
After we perform classification of our samples, we try to find a graspable region with relatively high classification confidence score. 
Our grasp optimization builds on the particle filtering work proposed by Dellaert et al.~\cite{Dellaert_MCL}, which is based on sequential Bayesian filter:
 \begin{equation}
　　\label{eq:bf}
　　　Bel(q_t) \propto p(z_t|q_t)\sum_{j}p(q_t^{(j)}|q_{t-1}^{(j)})Bel(q_{t-1}^{(j)})
\end{equation}
where the weighted particles $\{q_t^{(j)},w_t^{(j)}\}_{j=1}^{n}$ represent the sampled six-DoF grasp poses with confidence score given by classifier. The initial hypothesis of particles $q_t^{(j)}$ are uniformly generated in the 3D workspace with the identical weights. For each hypothesis, we extract the grasp features and compute the weight $w_t^{(j)}$ by normalizing the confidence score output by classifier. Importance sampling is then performed with resampling process to concatenate grasp hypothesis to high weights region. 
In our case, we don't have actual action between two states, instead, we model the state transition in action model as zero-mean Gaussian noise over $SE(3)$. In other words, after we obtain resampled grasp poses (particles), we diffuse the particles by adding Gaussian noise over $(x,y,z,row,pitch,yaw)$ to generate the new set of particles. Our convergence criterion is a fixed number of iterations.

\section{Results}

\subsection{Experimental Setup}
To evaluate \glassloc{}, we ran a series of experiments with a first generation Lytro camera and a Michigan Progress Fetch robot. The Lytro camera is mounted on the wrist of the robot and triggered by on-chip Wi-Fi to take images. In the meantime, the robot will record the camera view pose based on the current transformation from robot base to the camera. 
The Lytro camera intrinsic calibration and distortion correction is conducted using the toolbox created by Bok et al.~\cite{bok2017geometric}. The raw light field image is then decomposed into $9\times9$ sub-aperture images with resolution of $328\times328$ pixels. The boundary sub-aperture images usually have strong color noise because of the lens edge affect. In our implementation, we only keep $7\times7$ sub-aperture images and for each image. For each image,  we crop 4 pixels at the margin.

\begin{table}
\begin{center}
\begin{tabular}{ ccc } 
 \hline
 Object & Trials & Success Rate \\ 
 \hline
 Toothbrush Holder & 60 & 0.92 \\ 
 Wine Cup & 50 & 0.82  \\ 
 Short Cup & 40 & 0.65 \\ 
 Tall Cup & 40 & 0.88 \\ 
 Spoon & 30 & 0.70 \\ 
 
 \hline
 Overall & 220 & 0.81 \\
\end{tabular}

\caption{Object-wise grasp performance}
\label{table:res_singleobj}
\end{center}
\end{table}

\newcommand{\subfigwidth}{0.22}
\begin{figure*}

\subfloat{
\includegraphics[width=\subfigwidth\linewidth]{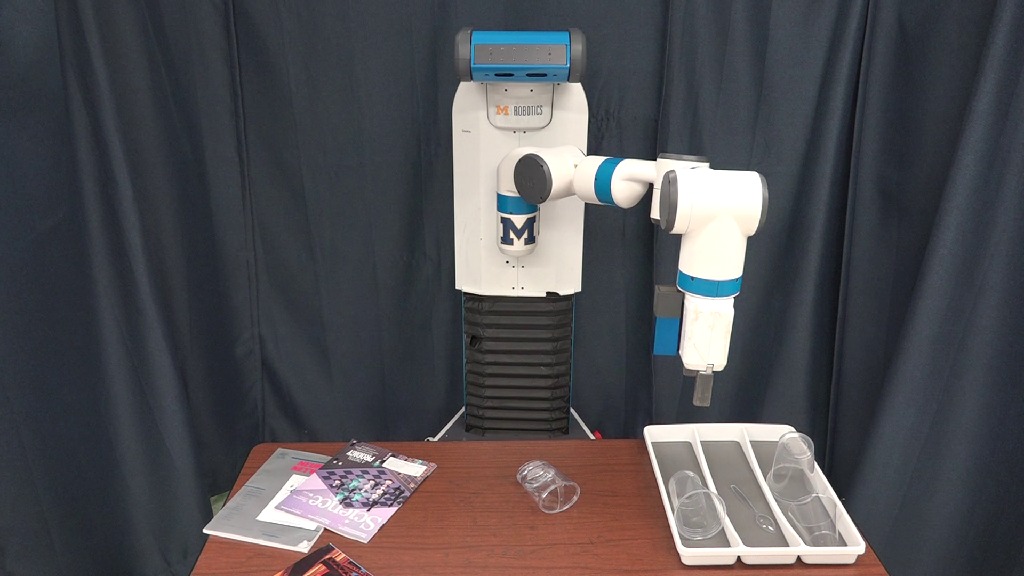}
}
\subfloat{
\includegraphics[width=\subfigwidth\linewidth]{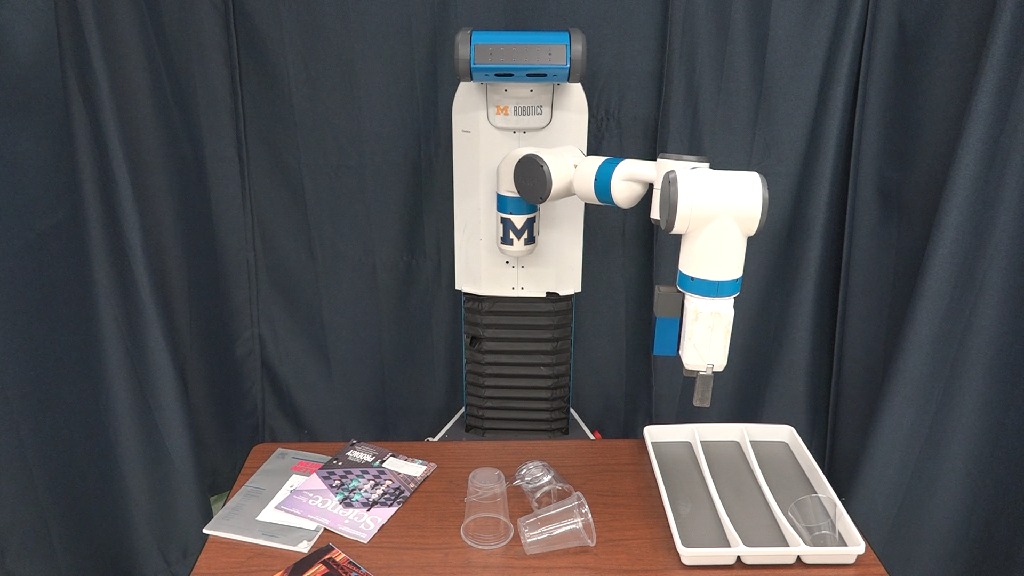}
}
\subfloat{
\includegraphics[width=\subfigwidth\linewidth]{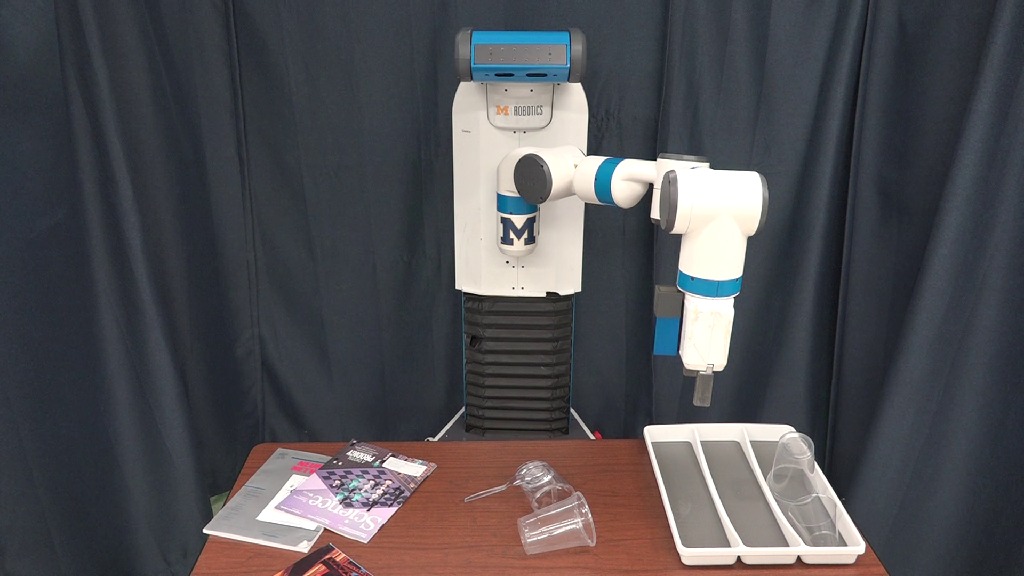}
}
\subfloat{
\includegraphics[width=\subfigwidth\linewidth]{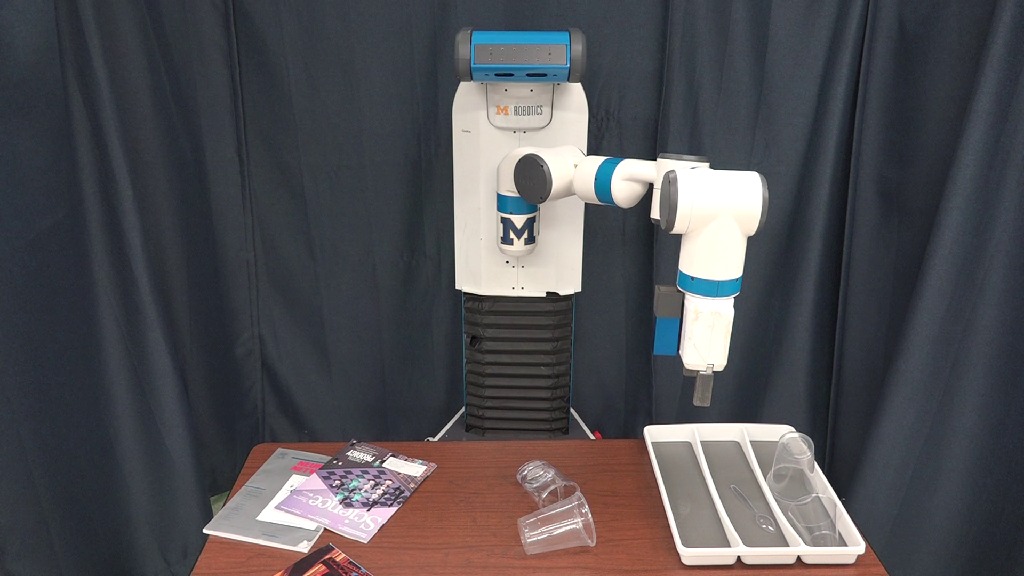}
}

\subfloat{
\includegraphics[width=\subfigwidth\linewidth]{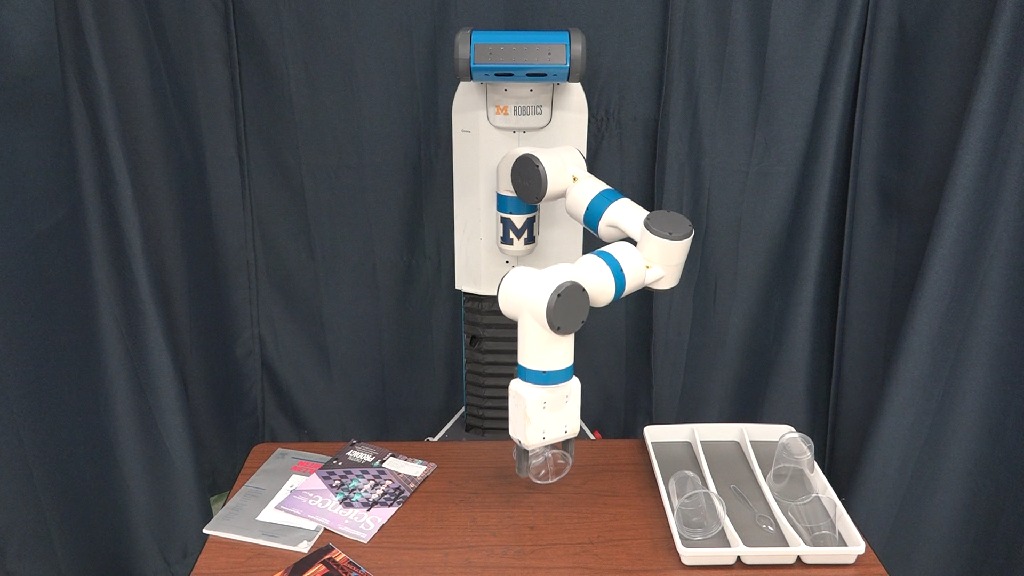}
}
\subfloat{
\includegraphics[width=\subfigwidth\linewidth]{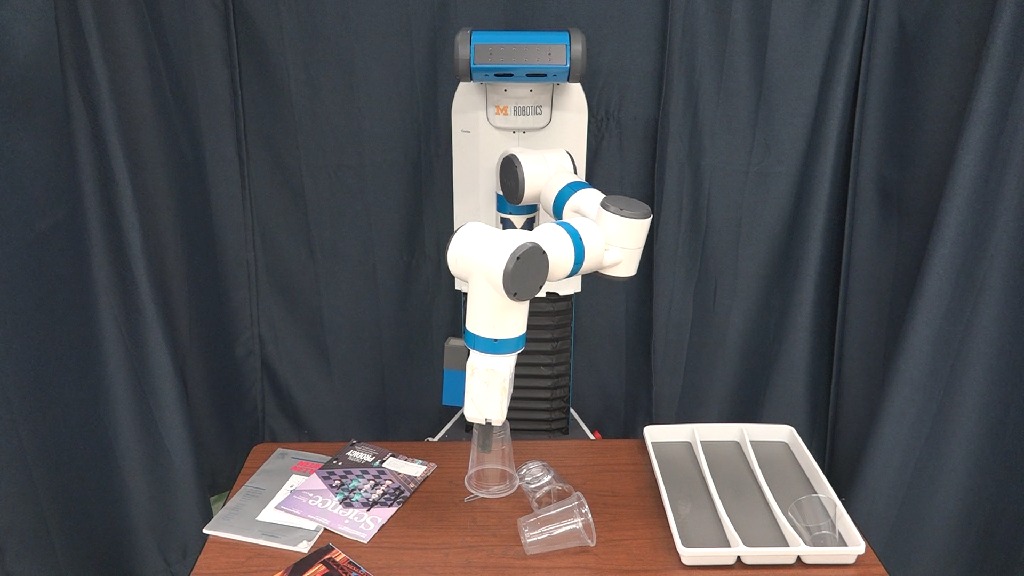}
}
\subfloat{
\includegraphics[width=\subfigwidth\linewidth]{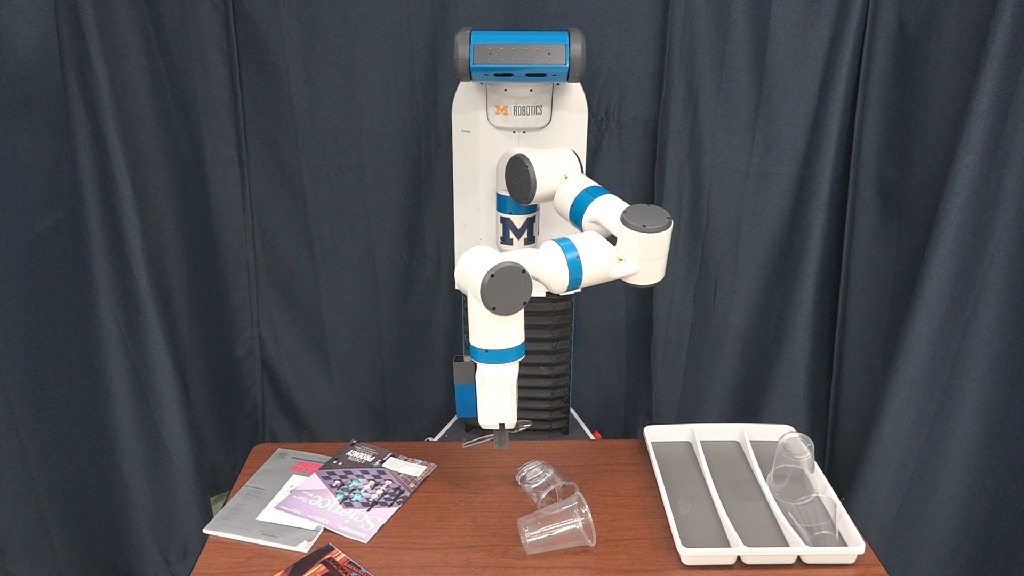}
}
\subfloat{
\includegraphics[width=\subfigwidth\linewidth]{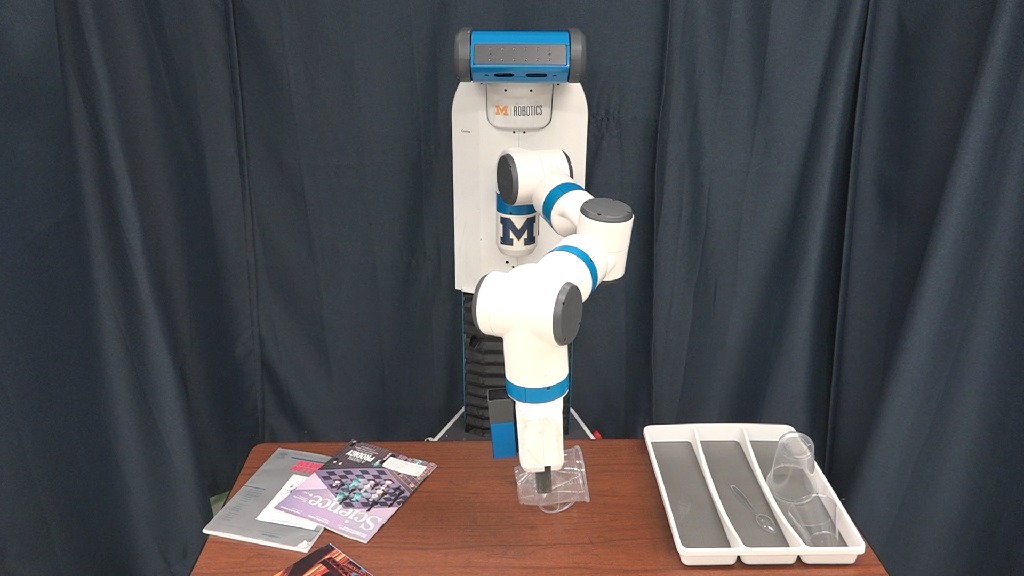}
}

\subfloat{
\includegraphics[width=\subfigwidth\linewidth]{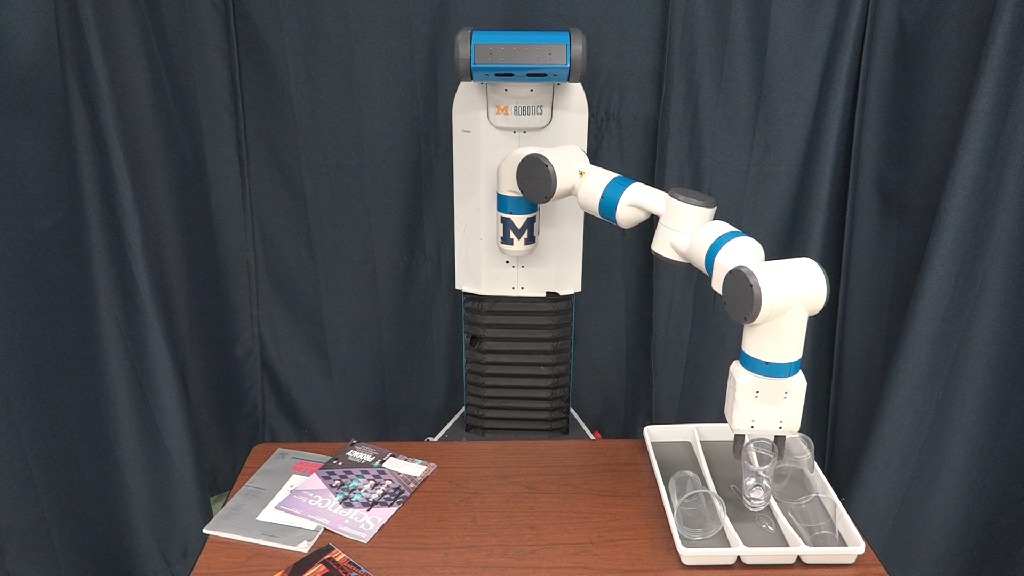}
}
\subfloat{
\includegraphics[width=\subfigwidth\linewidth]{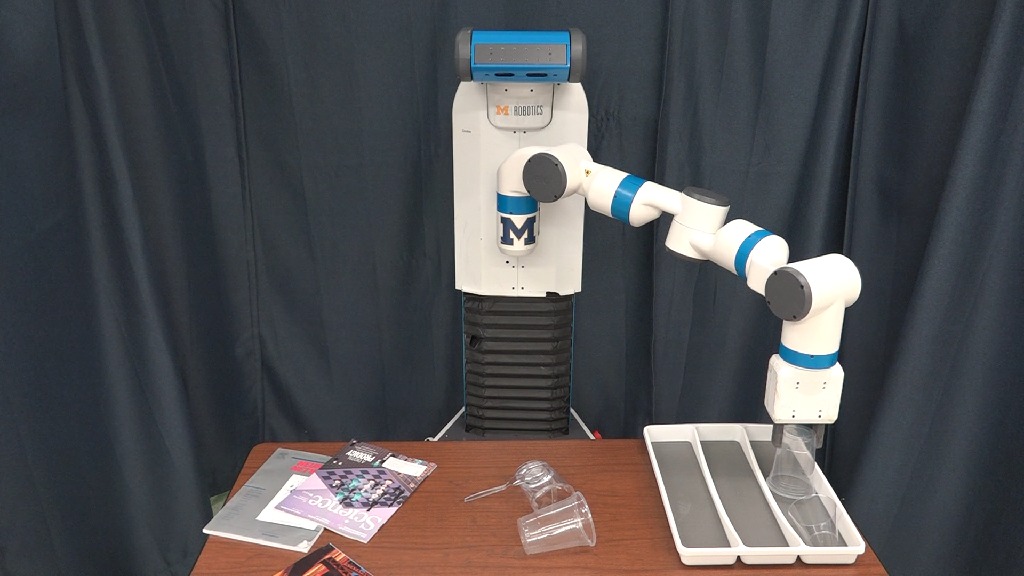}
}
\subfloat{
\includegraphics[width=\subfigwidth\linewidth]{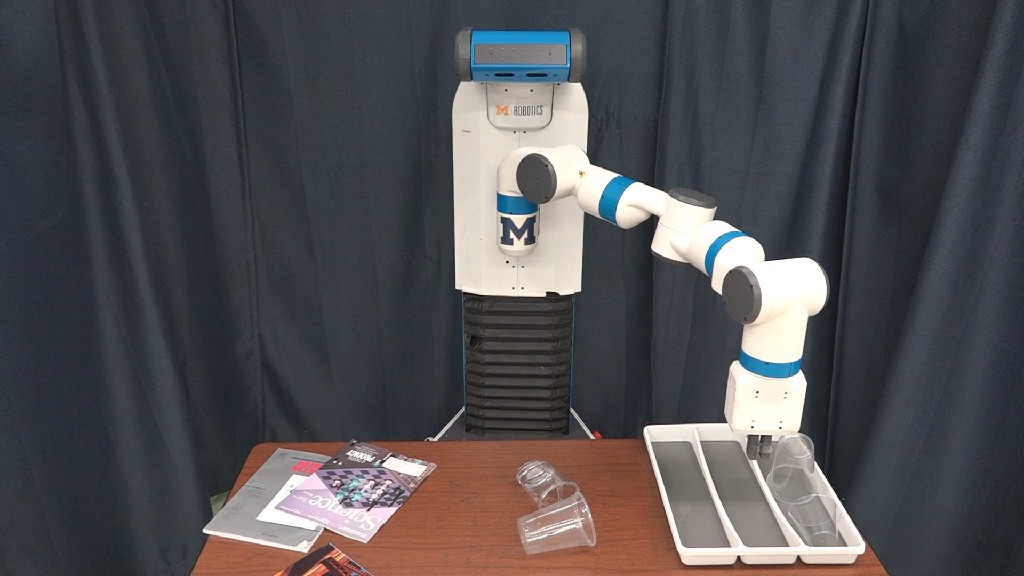}
}
\subfloat{
\includegraphics[width=\subfigwidth\linewidth]{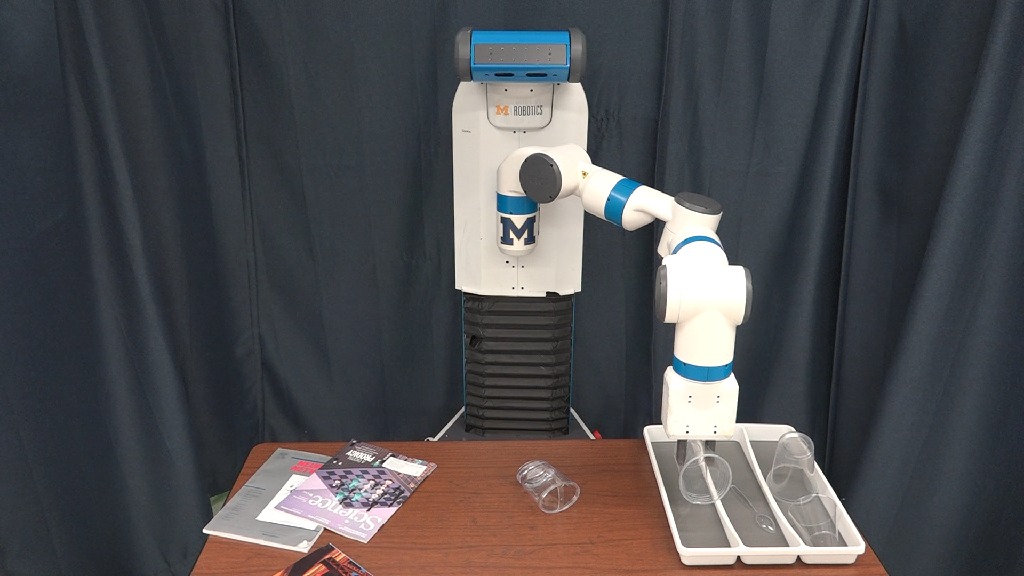}
}
\caption{The robot successfully picks and places all transparent objects in scene (g). Each column shows the pick and place action over one object in the scene. }
\label{fig:exp_example}
\end{figure*}


We use two objects to construct our training samples: wine cup and short cup (Figure~\ref{fig:train_test}). We generate approximate 10k positive grasp samples and 15k negative grasp samples from 50 scenes containing one or more object instances. For each grasp sample, we extract corresponding graspable volume from DLV with actual size $0.10 \times 0.10 \times 0.06\;(meters)$ and grid density $100\times 100 \times 60\;(points)$. We further extract gray-scale image features and resize them into $100 \times 100$. Features are concatenated into nine channels and trained on LeNet structure. We keep the default structure and parameter settings of LeNet implementation in Tensorflow except the number of nodes in the output layer (2 in our case). 

The DLV construction algorithm is implemented in MATLAB with parallel computing. A DLV is sampled in a $1.0 \times 1.0 \times 1.0\;(meters)$ box with grid density at $1000 \times 1000 \times 1000\;(points)$. 

In grasp search step, we use 100 particles with 100 iterations in our experiment. The covariance for diffusing grasp pose after each filtering iteration is set to $10^{-4}\;(meter^2)$ and $0.03\;(rad^2)$ for translation and rotation respectively. 

Our implementation takes 2 minutes per view to extract sub-aperture images and 10 minutes to construct DLV on an
unoptimized MATLAB code. The light field image decoding and ray corresponding are the current bottlenecks.
\subsection{Evaluation}
We evaluate our \glassloc{} manipulation pipeline on eight transparent clutter scenes as shown in Figure~\ref{fig:test_scenes}. In each scene, the number of objects ranges from two to four with different pose configurations.
For each manipulation run, light field images are taken from two camera poses to construct DLV.
After particle filtering reaches the convergence criterion, we randomly select one grasp pose and send it to the execution module. Our robot motion planning and execution module is built on TRAC-IK~\cite{beeson2015trac} and MoveIt!~\cite{sucan2013moveit}. For each scene, we perform 10 manipulation runs. We will terminate one run whenever all objects are successfully picked or the number of manipulation trials exceed the number of objects.

The manipulation results of each scene are established in Table~\ref{table:results}. Object grasp percentage is calculated based on how many objects have been successfully picked over the total number of objects that should be picked in all runs of a scene. We also show the pick success rate for each object in Table~\ref{table:res_singleobj}. 

Table~\ref{table:results} shows that the object grasp percentage is over 75\% in most of the scenes. Our \glassloc{} algorithm can generate enough reliable grasp poses based on our DLV constructed from light field observations in complex scenes where four transparent objects are randomly cluttered. The grasp percentages of these two scenes are 100\% and 85\% respectively.

Notably, our overall grasp success rate is 81\% for the transparent cluttered environments in 220 grasps. During our experiment, we find that the short cup has the lowest grasp success rate. In most cases, it was squeezed and then slipped out from the gripper. The reason is two fold: one is that the surface of the short cup is sharply tilted, which prevents the robot from performing force closure grasping, the other is that the parallel jaw gripper hasn't been equipped with force sensors and is likely to squeeze the cup.
\ignore{In consequence, the configuration of the short cup in scene(f) makes it difficult for the robot to successfully grasp. In comparison, the success rate of the toothbrush holder is high because the robot can perform force closure grasping on it. The tangent planes of its surface at the contact points are almost parallel, which makes the holder remain static balance at grasping.}


\section{Conclusion}
In this paper, we have contributed the \glassloc{} algorithm for robot manipulation in transparent clutter. We use multi-view light field observations to construct the Depth Likelihood Volume as a plenoptic descriptor to characterize the environments with multiple transparent objects. We show that by our algorithm, the robot is able to perform accurate grasping in tabletop transparent cluttered environments.


\balance



\bibliographystyle{abbrv}
\bibliography{big}

\end{document}